\documentclass[]{article}

 \usepackage[final]{nips_2018}

\usepackage[utf8]{inputenc} 
\usepackage[T1]{fontenc}    
\usepackage{hyperref}       
\usepackage{url}            
\usepackage{booktabs}       
\usepackage{amsfonts}       
\usepackage{nicefrac}       
\usepackage{microtype}      
\usepackage{graphicx}

\usepackage[disable]{todonotes}

\newcommand{\chugo}[1]{\todo[color=red!60,inline]{H:#1}}

\title{Continual State Representation Learning for Reinforcement Learning using Generative Replay}

\author{Hugo Caselles-Dupr\'e\textsuperscript{1,2}, Michael Garcia-Ortiz\textsuperscript{2}, David Filliat\textsuperscript{1} \\
Flowers Laboratory (ENSTA ParisTech \& INRIA)\textsuperscript{1}, AI Lab (Softbank Robotics Europe)\textsuperscript{2}\\
\href{mailto:caselles@ensta.fr}{caselles@ensta.fr}, \href{mailto:mgarciaortiz@softbankrobotics.com}{mgarciaortiz@softbankrobotics.com}, \href{mailto:david.filliat@ensta.fr}{david.filliat@ensta.fr}}

\begin{document}

\maketitle
\begin{abstract}

We consider the problem of building a state representation model in a continual fashion. As the environment changes, the aim is to efficiently compress the sensory state's information without losing past knowledge. The learned features are then fed to a Reinforcement Learning algorithm to learn a policy. 
We propose to use Variational Auto-Encoders for state representation, and Generative Replay, i.e. the use of generated samples, to maintain past knowledge. We also provide a general and statistically sound method for automatic environment change detection. 
Our method provides efficient state representation as well as forward transfer, and avoids catastrophic forgetting. The resulting model is capable of incrementally learning information without using past data and with a bounded system size.

\end{abstract}

\section{Introduction}

Building agents capable of learning over extended periods of time in the real world is a long standing challenge of Reinforcement Learning (RL) research, with direct applications in Robotics. Such an agent should be able to continually learn about its environment. This involves building a model of its surroundings, with visuals features, and continually updating this model as its life progresses and the environment evolves. It is common to use state features to learn to solve tasks using RL algorithms, a field known as State Representation Learning (SRL) \citep{lesort2018state}. However, since these models are often neural networks trained using stochastic gradient descent (or any variant), they forget past knowledge when the training data distribution changes, an infamous problem called catastrophic forgetting \citep{french1999catastrophic}. 
We propose to use a class of generative models often used for SRL, Variational Auto-Encoders (VAEs), combined to a Continual Learning (CL) approach to overcome this issue. 
Previous work have shown that VAEs can learn continually using Generative Replay \citep{anonymous2018generative}. The method uses the generative ability of VAEs to remember and re-use past knowledge. With the goal of an autonomous agent in mind, we complement our approach with a general method for automatic detection of environment change.

We test our approach on a 2-D first-person environment with coherent physics, in a scenario where the environment changes. We measure the ability of our method to retain past knowledge, using reconstruction error. We also test whether the learned features provide efficient and high-performing RL training. Our results show that our approach avoids catastrophic forgetting and has a form of forward transfer, i.e. the ability to better solve a task using previous knowledge. 
Additionally, it respects important desiderata: no access to past data and bounded system size. Finally, the user needs not to specify environment changes. 

\section{Background}


Variational Auto-Encoders \citep{kingma2013auto} are a particular kind of auto-encoders that learn to map their input into a continuous latent representation. The loss function is composed of two terms: the reconstruction error, and a regularization term that constraints the posterior distribution.
When trained on an agent's inputs, they provide a compressed vectorial representation of what the agent experience. They have recently shown promising results as State Representation models for RL \citep{ha2018recurrent, wayne2018unsupervised}. 

While most work in CL is focused on discriminative models, there is a recent interest in CL approaches for generative models \citep{achille2018life, nguyen2017variational}. Proposed methods often rely on using generated samples to avoid forgetting \citep{wu2018memory, anonymous2018generative}. This technique, which we use in this paper, is termed Generative Replay. To the best of our knowledge, we found no previous work on applying these CL approaches to State Representation Learning for RL. The closest work to this paper is DARLA \citep{higgins2017darla}, which circumvent the problem of catastrophic forgetting by learning disantangled representations with a specific VAE architecture. Their approach learns features that are robust to minimal modifications of the environment, while ours continually update features as environment changes are detected. Our work is also close to Transfer Learning in a RL context \citep{taylor2009transfer}.

\begin{figure}
    \centering
    \includegraphics[scale=0.22]{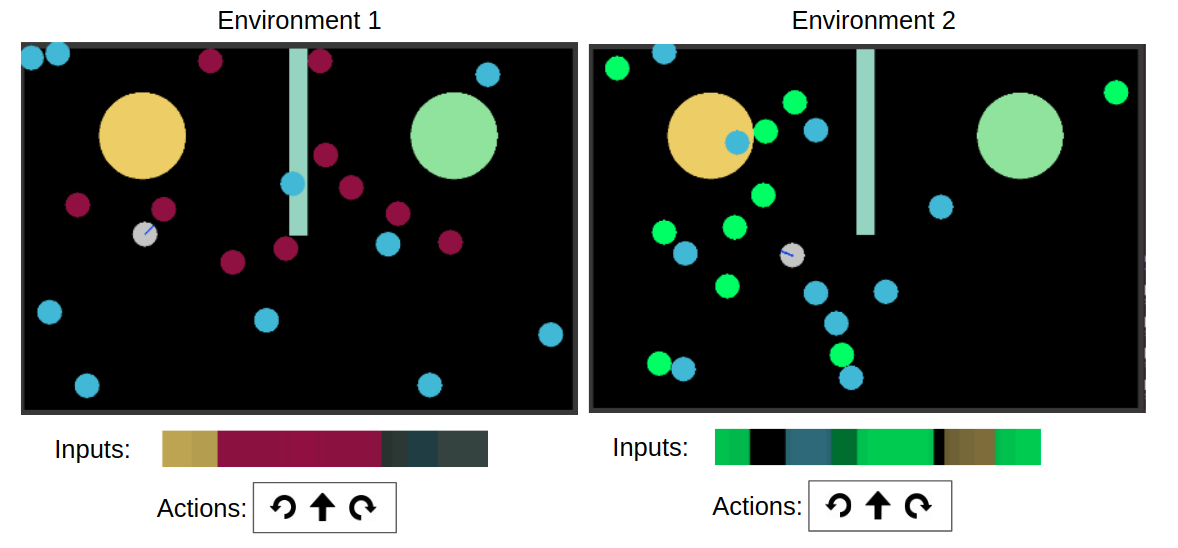}
    \caption{The two environments considered in this paper.}
    \label{fig:env_explained}
\end{figure}


\section{Continual State Representation Learning with Generative Replay}


\subsection{Learning continually with Generative Replay}
The considered problem is twofold: 1) build a State Representation model 2) which can continually learn, i.e. without forgetting. For 1), we propose to use VAEs' encoding property 
to handle the State Representation. For 2), we propose to use the generative ability of VAEs to generate states of previously seen environments to avoid forgetting, a technique known as Generative Replay \citep{shin2017continual}. In our case, we encode the sensory state received by the agent.

Once the environment changes, we generate states using the latent space of the VAE trained on previous environments, and add those generated samples to states collected in the new environment. Then, we train the VAE on the joint data. This option is scalable (bounded system size) and does not re-use past data. The environment changes are detected automatically, as described thereafter.

\subsection{Automatic environment change detection}
Since we aim at constructing an autonomous agent, we complement the proposed method with automatic detection of changes in the environment. The method is based on a statistical test on VAE reconstruction error distribution. We use Welch's $t$-test \citep{welch1947generalization} to test the hypothesis $\mathcal{H}_0$ that two means $x_1, x_2$ of randomly sampled VAE reconstruction errors have equal mean. We choose this test over the standard $t$-test because we do not have reasons to assume that the two samples' variance are equal. The statistic $t$ is, under $\mathcal{H}_0$, distributed as a Student distribution with a number of degrees of freedom $\nu$ that can be approximated using the Welch–Satterthwaite equation: 
\begin{equation}
    t=\frac{x_1 - x_2}{\sqrt{(s_{1}^{2}+s_{2}^{2})/N}}, \quad \nu \approx\frac{(N-1)(s_{1}^{2}+s_{2}^{2})^2}{(s_{1}^{4}+s_{2}^{4})},
\end{equation}
with N being the number of samples (assumed equal for both samples), and $s_1, s_2$ being the empirical standard deviations of samples 1 and 2, respectively.



Using a statistical test is preferable to using a threshold. The threshold is based on an arbitrary scale which depends on the considered environment. On the contrary, the test is a more general approach based on statistical principles and thus agnostic to scales. Additionally, we choose to use VAE reconstruction error distribution over the actual state distribution because we consider changes in the environment that require the VAE to be updated. For instance, if we add an already existing obstacle to the environment, the state distribution shifts, while the VAE reconstruction error distribution does not change because the VAE need not to be updated. The test is lightweight and can thus be continually performed.

\section{Experiments}
\subsection{Environments and methods}


Our experiments use two environments developped in Flatland \citep{caselles2018flatland}, see Fig. \ref{fig:env_explained}. Most elements are identical between the two worlds: both are rooms of the same size with 3 fixed obstacles, 10 randomly placed round blue obstacles and 10 randomly placed round edible items. The only variation is the color of the edibles items. The navigation task consists in collecting as many edible items as possible in $500$ timesteps. The input is a 1-D image corresponding to what the agent sees in front of it. The navigation task, inputs and actions are detailed in Fig. \ref{fig:env_explained} and Appendix \ref{app:setup}.


For the State Representation model, we train the VAE with a modified version of KL annealing \citep{bowman2015generating}, see Appendix \ref{app:repro} for details. For our RL experiments, we use the Proximal Policy Optimization (PPO) algorithm \citep{schulman2017proximal} of SRL Toolbox \citep{raffin2018s}. We selected this method as it is a state-of-the-art policy gradient method, commonly used and robust to hyperparameter configurations. Implementation details are provided in Appendix \ref{app:repro}.

\chugo{Parler des comparaisons.}

\subsection{Evaluation}
First, we evaluate whether the VAE successfully learned to reconstruct states, generate realistic states using sampling in the latent space and avoid catastrophic forgetting. We use visualization to measure performance. We also use the Mean Squared Error (MSE) to evaluate reconstruction quality and catastrophic forgetting.
Then, we evaluate the VAE ability to provide a state representation that enables efficient and high-performing RL training, by measuring the performance of an RL agent using VAE features as input. The aim of SRL here is to have features that enable efficient behaviour learning hence the need to evaluate with RL. We compare our approach, Generative Replay, to learning with raw inputs (no information compression) and to Fine-tuning, where the VAE is naively fine-tuned on the second environment.

\begin{figure}[h!]
    \centering
    \includegraphics[scale=0.25]{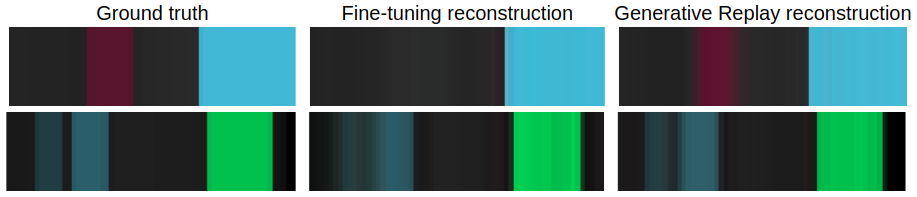}
    \caption{Reconstruction comparison: VAE fine-tuned on environment 2 against Generative Replay.}
    \label{fig:reconstruction_forgetting}
\end{figure}

\section{Results}
\label{sec:results}

\chugo{Talk about KL annealing here, and beta VAE. Dire de DARLA ne marche pas. VIDEOS}

\hspace{0.5cm} \textbf{Automatic detection of environment change}: Our proposed automatic detection method tests whether a VAE trained on the first environment has a mean reconstruction error statistically different between states of environments 1 and 2. For that, we use a Welch's $t$-test to compare two batches of mean VAE reconstruction error, computed on randomly collected states. The null hypothesis is rejected if the $p$-value is greater than $0.01$. We repeat this experiment $5000$ times. The test is 100\% successful when it should detect an environment change, and 99.5\% successful when it should not detect a change. Any chosen critical $p$-value between $0.05$ and $0.0001$ provide similar results. Details for this experiment are provided in Appendix \ref{app:auto_detect}. 

\begin{figure}[h!]
    \centering
    \includegraphics[scale=0.3]{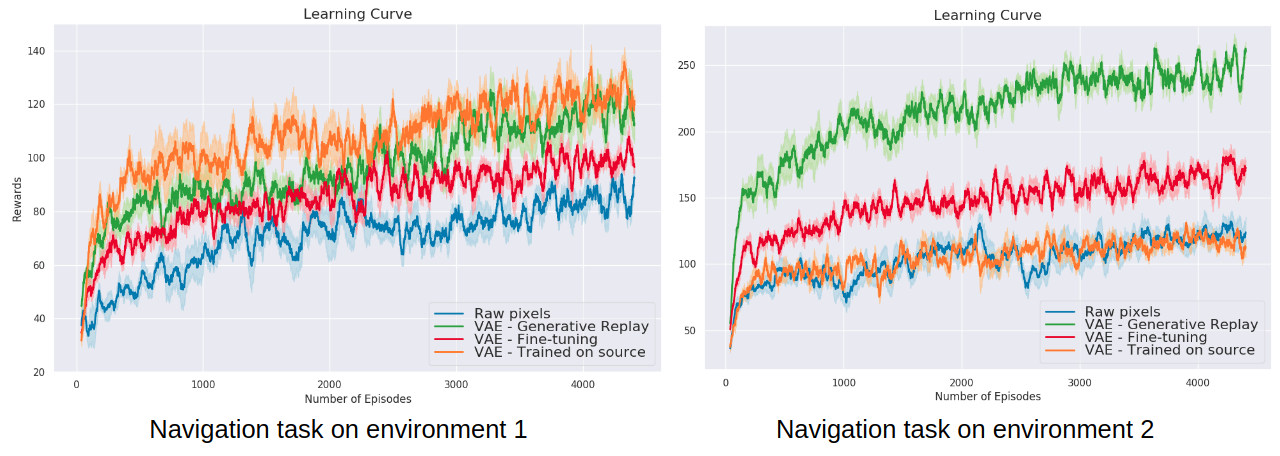}
    \caption{Mean reward and standard error over 5 runs of RL evaluation using PPO with different inputs. Fine-tuning and Generative Replay models are trained sequentially on the first and second environment, and then used to train a policy for both tasks.}
    \label{fig:rl_eval}
\end{figure}



\hspace{0.5cm} \textbf{Reconstruction evaluation}: We present in Fig. \ref{fig:reconstruction_forgetting} a qualitative evaluation of Fine-tuning and Generative Replay methods on VAE. The model is sequentially trained on the first and then the second environment. Naive Fine-tuning of VAE on the second environment leads to forgetting of the ability to reconstruct states from the first environment. On the contrary, Generative Replay successfully avoids this problem and the resulting VAE is able to properly reconstruct all elements of both environments, hence successful continual learning. Quantitatively, we observe the same results, see Table \ref{tab:mse_reconstruction}. The MSE of reconstruction over $500$ samples is similar for the two method on environment 2, whereas Generative Replay is one order of magnitude better than Fine-tuning on environment 1.
\begin{table}[!htb]
\centering
  \caption{Mean Squared Error (MSE) of reconstruction.}
  \label{tab:mse_reconstruction}
  \scalebox{0.8}{
  \begin{tabular}{ccc}
    \hline 
    Strategy& MSE (environment 1)& MSE (environment 2)\\ 
    \hline
    Fine-tuning  & $1.3e10^{-3}$   &  $9.3e10^{-4}$ \\
    Generative Replay& $3.3e10^{-4}$ & $6.4e10^{-4}$ \\
    \hline
\end{tabular}}
\end{table}


\hspace{0.5cm} \textbf{RL evaluation}: Learning curves are presented in Fig. \ref{fig:rl_eval}, and final performance are reported in Table \ref{tab:final_rl_perf} of Appendix \ref{app:add_results}. We first verify that using State Representation instead of directly using the raw states is superior in terms of final performance and sample efficiency. Also, it is easier to obtain better rewards on environment 2. Contrary to task $1$, the color of the edible items is focused on one color channel on task $2$, which supposedly facilitates policy learning. We performed an additional experiment on a different color transition which verified that the results persisted, see Appendix \ref{app:add_exp}. 

We observe that VAE's features supports a form of zero-shot transfer. Indeed, using features of a VAE trained on the first environment to learn to solve task $2$ is remarkably efficient. Incidentally, using features of a VAE fine-tuned on the second environment to learn to solve task $1$ is also quite satisfactory, even if we previously showed that the model has catastrophically forgotten how to reconstruct states of the first environment. Yet, there is still significant gap in performance compared to using a model trained on the first environment, so forgetting is still apparent. 
Our method, Generative Replay, completes this gap as it does not forget how to encode states from the first environment. Additionally, we observe on task $2$ that it largely outperforms all other methods. We infer that learning on both environments produces representations that are more suited for policy learning, a form of forward transfer. Videos of learned policies using VAE with Generative Replay are provided here: \url{https://drive.google.com/open?id=1ilkgWtDqL-F2ZZtiBprvrde2x6kC3r9i}.



\section{Conclusion and future work}

We presented a VAE-based State Representation model capable of learning continually, as the environment changes and with a fixed-size system. 
Our method automatically detects changes and relies on using generated samples of previous environments. Learned features allow efficient and high-performing RL. 
For future work, we will be experimenting on a more challenging setting, e.g randomly generated mazes, see Fig. \ref{fig:futurework} in Appendix \ref{app:future_work} and with different environment changes. Another direction is to extend this method to non-discrete environment changes.

\bibliographystyle{ACM-Reference-Format}
\bibliography{nips_2018} 

\appendix

\section{Experimental details}

\subsection{Environments and task for RL experiments}
\label{app:setup}

All colors in the environment are fixed. In environment 1, the edible items are red, and they are green in environment 2. The agent receives $+10$ reward for each collected edible item, which is its only source of reward. The goal of the agent is to find a strategy to get a maximum reward in a fixed amount of timesteps ($500$). The agent's action space is discrete and composed of 3 actions: \textit{move forward}, \textit{rotate left} and \textit{rotate right}. It receives as input a 1-D image corresponding to what the agent sees in front of it. The agent always spawns at the same location.

\subsection{Automatic detection of environment change}
\label{app:auto_detect}

We compare two batches of reconstruction error of a VAE trained on environment 1. There are two cases:

\begin{itemize}
    \item Environment 1 vs environment 1: if the two batches are reconstruction errors of states collected on the same environment, the method should not detect a change.
    \item Environment 1 vs environment 2: if the two batches are reconstruction errors of states collected on different environments, the method should detect a change.
\end{itemize}

We construct batches as follows: we compute the mean reconstruction error of randomly collected states over $20$ episode, and we add the value to the batch. We repeat this process $10$ times, so that batches are 10-dimensional vectors.

Then the statistical test is computed by comparing two batches, computing the p-value $p$ and comparing it to the reference value of $0.01$. If $p>0.01$ then there is no detected change, otherwise there is. 

The method is 100\% successful when it should detect an environment change, and 99.5\% successful when it should not detect a change.

\subsection{Architectures and hyperparameters}
\label{app:repro}

The size of the 1-D image inputs is $(64,3)$. We fix the size of the latent representation to $64$. We use three 1-D convolutional layers followed by one fully connected layer for the encoder and decoder. We choose the Rectified Linear Unit activation function. The batch size is $128$. 

We describe here our version of KL annealing used in our experiments. The initial KL annealing method puts a weight of $0$ on the KL term and smoothly increase it to $1$. We found standard training (without KL annealing) and this method ineffective: the models were unable to reconstruct inputs properly. Our version initializes training with a weight on the KL term of $1$, as in standard training, then smoothly reduces it to $0$ until training is stopped when the reconstruction error does not improve. This inverse annealing scheme allowed successful training, as presented in Section \ref{sec:results}. The annealing parameter is $0.9995$. At each new batch, we update the coefficient in front of the KL term by this annealing parameter. 

For training, we collect states from $1000$ episodes of $500$ timesteps each, of a random policy. Hence, the training set is composed of $500000$ states. 

For Generative Replay, we sample the same number of states from the latent space of the VAE trained on environment 1, and append those states to the $500000$ states collected randomly in environment 2. To stop training, we use early stopping on the reconstruction error with a threshold of $0.001$. If the reconstruction error does not improve more than the threshold over $5$ epochs then we stop training and select the least recent model. 

We use the standard version of PPO in the S-RL Toolbox implementation \citep{raffin2018s}, see their code for details. The method learns a policy such that after an update, the next policy is not too different from the previous policy. This is done using clipping to avoid a too large update. The policy is a Multi-Layer Perceptron with Relu activations. 

\subsection{Additional results}
\label{app:add_results}

The final performance of the RL evaluation is presented in Table \ref{tab:final_rl_perf}.

\begin{table}[!h]
\centering

  \caption{Mean final performance of RL evaluation.}
  \label{tab:final_rl_perf}

  \begin{tabular}{ccc}
    \hline 
    Inputs& Task 1& Task 2\\ 
    \hline
    Raw pixels & $92.30 \pm 5.8$   &  $123.95 \pm 25.6$ \\
    VAE - Trained on source  & $121.25 \pm 5.3$   &  $111.75 \pm 11.9$ \\
    VAE - Fine-tuning  & $96.55 \pm 5.1$   &  $172.5 \pm 11.5$ \\
    VAE - Generative Replay  &  $112.85 \pm 13.2$  & $256.95 \pm 10.3$ \\
    \hline
\end{tabular}
\end{table}

\section{Additional experiment}
\label{app:add_exp}

\begin{figure}[h!]
    \centering
    \includegraphics[scale=0.4]{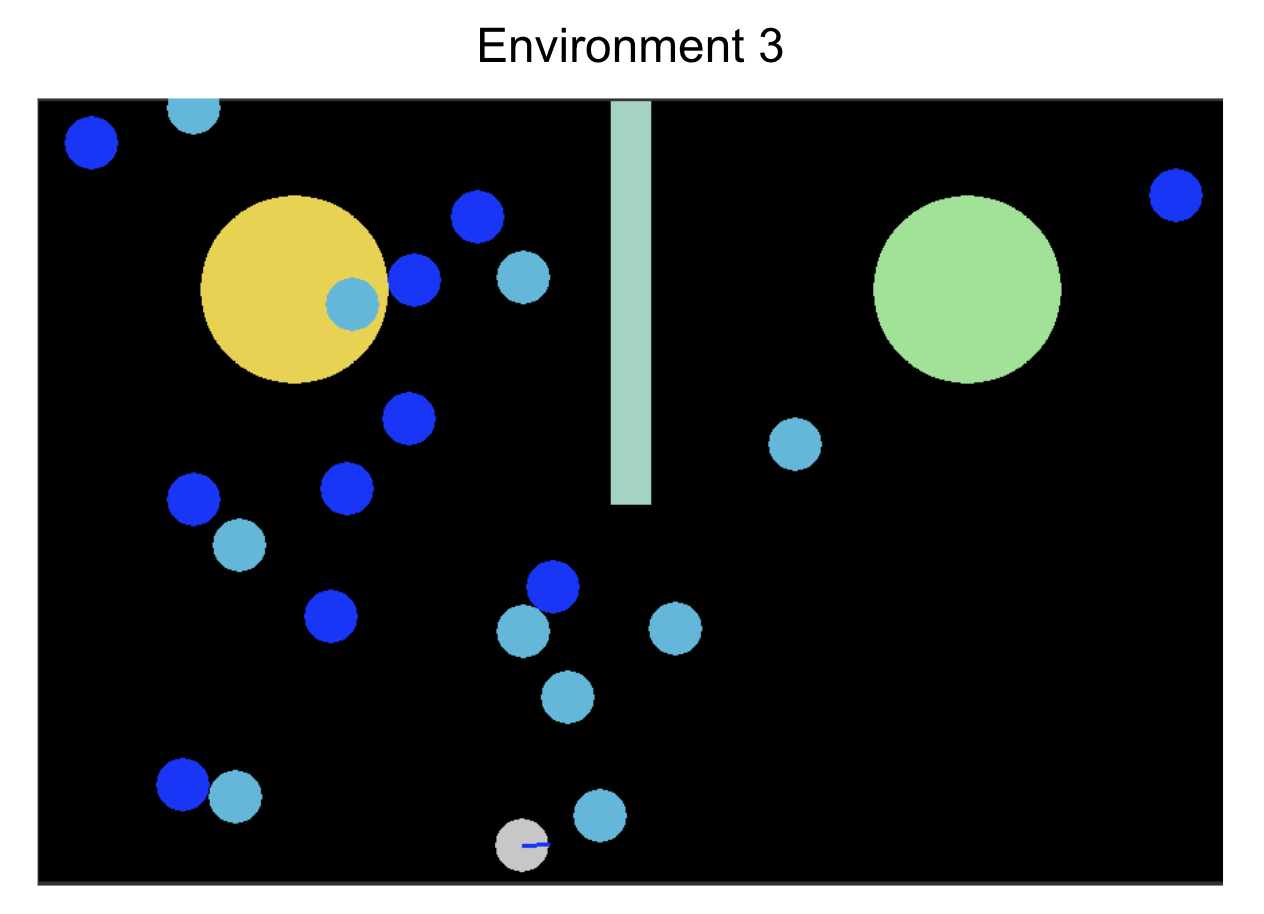}
    \caption{The third environment considered in this paper.}
    \label{fig:env3_explained}
\end{figure}

\begin{figure}[h!]
    \centering
    \includegraphics[scale=0.34]{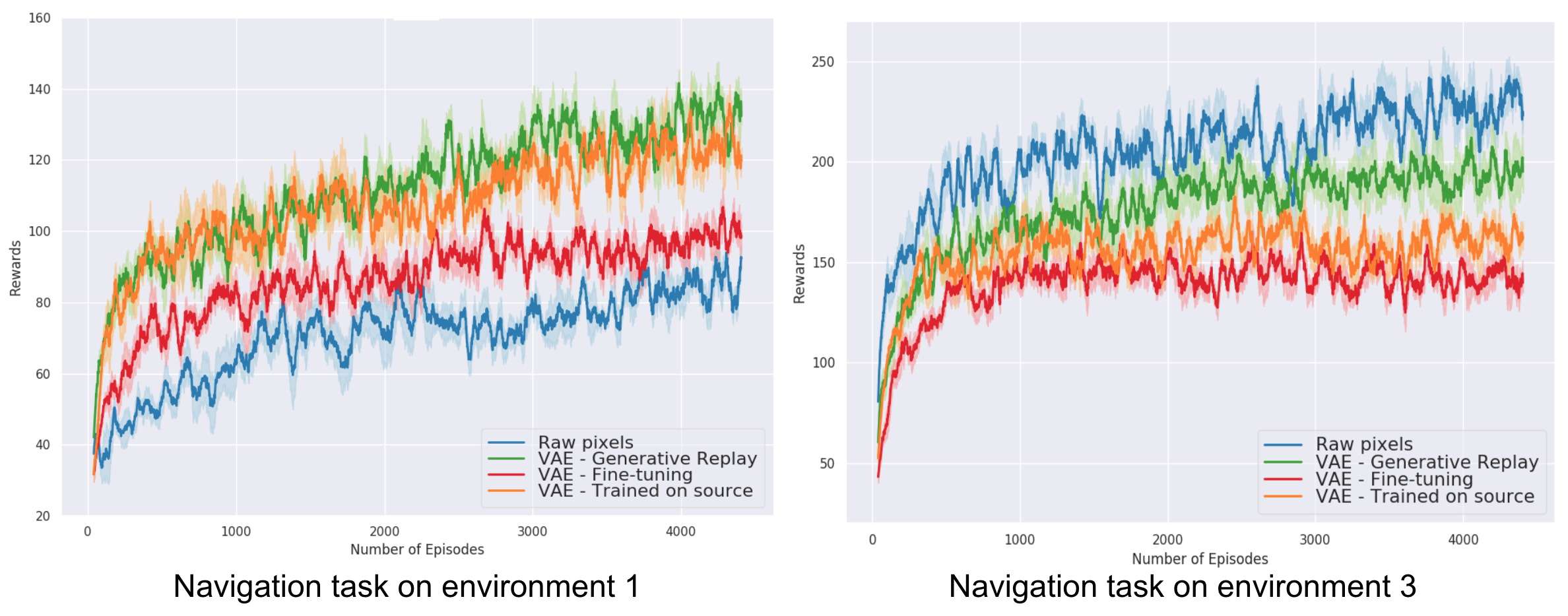}
    \caption{Mean reward and standard error over 5 runs of RL evaluation using PPO with different inputs. Fine-tuning and Generative Replay models are trained sequentially on the first and third environment, and then used to train a policy for both tasks.}
    \label{fig:rl_eval_v0_v3}
\end{figure}

We also provide in Fig. \ref{fig:rl_eval_v0_v3}, results for a similar experiment, but where the edible items in the second environment are blue rather than green. We call it environment 3, an illustration is provided in Fig. \ref{fig:env3_explained}. The purpose of this experiment is to show that the results are not specific to the color of the edible items.

The experiment confirms that: 

\begin{itemize}
    \item Fine-tuning catastrophically forgets.
    \item Generative replay outperforms all other VAE-based methods, and avoids forgetting.
    \item Generative Replay has a form of forward transfer.
    \item VAE features support a form of zero-shot transfer.
\end{itemize}

Thus, we can hypothesize that the positive results obtained in both experiments do not come from chance.

\section{Randomly generated mazes}
\label{app:future_work}

In this environment, a new maze is generated each $500$ episodes. Rooms and corridors change location, size and color. We plan on using this scenario to investigate how to continually learn without forgetting. This environment extends the considered setting in this paper. We will investigate how generative models can be used to learn a continual state representation under this more challenging setting.
\begin{figure}[h!]
    \centering
    \includegraphics[scale=0.7]{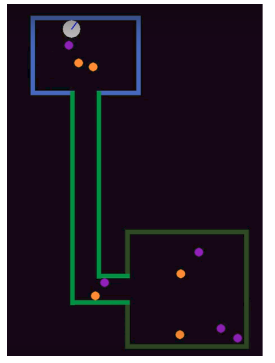}
    \caption{An illustration of randomly generated mazes.}
    \label{fig:futurework}
\end{figure}

\end{document}